# CAN A FACE TELL US ANYTHING ABOUT AN NBA PROSPECT? - A DEEP LEARNING APPROACH

**Andreas Gavros**
Aristotle University of Thessaloniki
Thessaloniki, Greece
agavros@arch.auth.gr

**Foteini Gavrou**
No affiliation
Kozani, Greece
foteiniga@gmail.com

## ABSTRACT

Statistical analysis and modeling is becoming increasingly popular for the world's leading organizations, especially for professional NBA teams. Sophisticated methods and models of sports talent evaluation have been created for this purpose. In this research, we present a different perspective from the dominant tactic of statistical data analysis. Based on a strategy that NBA teams have followed in the past, hiring human professionals, we deploy image analysis and Convolutional Neural Networks in an attempt to predict the career trajectory of newly drafted players from each draft class. We created a database consisting of about 1500 image data from players from every draft since 1990. We then divided the players into five different quality classes based on their expected NBA career. Next, we fit popular pre-trained image classification models in our data and conducted a series of tests in an attempt to create models that give reliable predictions of the rookie players' careers. The results of this study suggest that there is a potential correlation between facial characteristics and athletic talent, worth of further investigation.

## 1 Introduction

The inspiration for this paper was drawn from a series of articles published in 2014 [1, 2]. NBA professional teams, most notably the Milwaukee Bucks, worked with facial coding experts in an effort to evaluate the talent of the players who had declared for that year's draft class. So we made an attempt to create artificial intelligence models that would be able to mimic this human activity. Based on this logic, the goal of the assessment was to analyze certain facial characteristics to make an assessment of whether a young player is capable of standing at the NBA level and how good he will be.

Given the above, a legitimate question arises. Is there any truth or scientific basis to this analysis? Even further, what are the physiognomic characteristics that can provide information about aspects of a person's personality (if any) and more specifically their ability in a sport (basketball)? In this paper we will not attempt to answer this question. Instead we will attempt to construct Convolutional Neural Networks (CNNs) that will perform this task and evaluate them. CNNs are capable of extracting far more information than an image from a human observer [3]. Therefore, theoretically they will be able to outperform a human in assessing talent based on visual stimuli.

Foteini Gavrou provided all the image data through web-scraping and formatted the player images. Andreas Gavros, configured the database that was used to train the models. He developed and evaluated the CNN models. Lastly, he performed the literature review.

Obviously, we do not propose the methodology we follow in this study as a deterministic model to approach a player's potential talent or career. The main aim of this paper is more to explore one of the countless possibilities of CNNs and to provide an innovative approach to assess athletic talent, by professional athletic organizations. The operation of CNNs is still a problem to which the scientific community cannot provide a comprehensive answer [4]. Since we cannot yet ascertain how a CNN model "thinks", therefore, it is worth exploring their potential in projects that seemingly do not present a structured form, hence a seemingly explorable data pattern.

Can a face tell us anything about an NBA prospect?

The present work is divided in the following way:

1. **Introduction:** We provide the motivation for conducting the present study.
2. **Related Work:** We present important landmarks in the field of applied and computational statistics regarding professional sports research.
3. **Methodology:** We explain the methodology we used to create the database we used to train the pre-trained image classification models we deployed.
4. **Results:** We present the logic we followed to evaluate the models' predictions and the predictions of the best performing model.
5. **Conclusion:** We provide the central conclusions of this study.

## 2 Relevant Work

Since 1960 that sports analytics emerged as a field of sports analysis in basketball [5], statistical analysis of professional sports has made rapid progress. The monumental success of the Oakland Athletics' use of statistical methods in their decision-making process in the late 1990s [6] turned the attention of many sports organizations to the benefits of applied mathematics. In the start of 2000 decade professional teams, mainly from the NBA, like the Boston Celtics, made use of more sophisticated statistical methods, in their effort to evaluate a player's talent [7] as accurately as possible. The continuous boom of statistical methods that can be applied in sport has led over the last ten years to all sports organisations having a department dedicated exclusively to data analysis [8]. Now the explosion of the means available to collect and analyse data leads to the production of datasets that may no longer be usable, based on the statistical tools available [9].

As statistical analysis models prove their applicability to more and more areas of sport, more sophisticated analyses are being conducted. In the field of basketball, which is the subject of the present work, much research has been conducted on the evaluation of particular game strategies (the value of shooting threes in a game for example) [10] and the evaluation of in-game decisions by coaches [11]. The scope of DL has extended to areas of sport that have a huge impact, such as quantifying the influence of player injuries on a team's performance and investigating the potential correlation that may exist between these data and player performance [12].

Deep Learning is a field of computer science that has emerged as a prominent sports analysis tool that may have a significant impact in the performance of teams. CNNs have been used successfully, to process image input such as video and produce real-time data of a basketball game and measure is effects on the players and the coaches [13]. DL can be deployed to analyze which offensive play has the best performance in a basketball game and then optimize these systems to make them even more efficient [14]. Image analysis based on DL systems has found application in the investigation of the correlation between players' movement on the court and their efficiency. As it turned out DL models are capable of finding patterns in the offensive movements of players and suggesting optimal distances between players, as well as providing optimal positions on court for the initiation of certain plays [15].

Obviously a fairly significant portion of the relevant research is geared towards attempting to make predictions regarding the determination of talent in young players. The NBA's system for introducing new players to the league (draft classes) makes a successful draft selection very important to the success of an organization. Therefore, a lot of research has been carried out in this direction. Analyses have been conducted to explore the characteristics of draft prospects that influence decision-makers of NBA teams front offices to make their decisions [16]. Of course, teams are more interested in assessing a player's ability and whether he will be able to successfully adapt to the demanding environment of the NBA. On this question Machine Learning (ML) has some interesting answers to provide, as the analysis of statistical data from previous seasons can provide predictions about player development that cannot be made by an NBA team scout [17].

But here an important and obvious question arises: Can ML, CNN and DL algorithms outperform their human competitors in basketball talent evaluation? And if the answer to this question is yes, in what areas, in addition to those currently known, can these models be applied to provide even more critical information for NBA professional teams? In the latter question we will attempt to provide a new perspective by presenting the present research.

## 3 Methodology

The basis of the hypothesis we try to explore in this study is whether we can make one rough approximation of a rookie player's career, based solely on his facial features. We will then analyse the methodology we followed to produce the relevant forecasts.

## 3.1 Rationale of Study

The analysis was based exclusively on the data available from the 1990 season to the 2019 season, a period of 30 years, which is a relatively big amount of time regarding professional team sports. This choice was based on the following two factors: First, the availability of data. Image data of sufficient quality before 1990 is much more difficult to obtain. Second, the style of the game. Arguably, the game of basketball changed dramatically after the arrival of Michael Jordan [18]. Of course, for a few years after his selection in the 1984 draft [19], there were players who came from a much different era of the game. Therefore, we made the hypothesis that from 1990 onwards, the modern era of basketball began. During this era, we assume that there is a homogeneity in the characteristics of the players that are rated as most important by teams.

We also decided to work exclusively on the rookies (i.e. the newcomers to the NBA league). Finding the next great talent is a problem that all teams face every year. Based on the interest and importance of the problem, we decided to base our analysis on rookie talent evaluation.

## 3.2 Method of classifying players into talent categories

We decided to divide the players into five classes based on their *potential talent*. *Potential talent* is one of the key concepts of this study and refers to in the expected career of a player and not necessarily in the career he eventually had or will have. Furthermore, *Potential talent* refers to the career that a certain player was expected to have and not in the total numbers that a player put up in his career. To make this logic clearer we will present some real-life examples:

- The career stats for a player like Derrick Rose show a player who had some great seasons, but then had an overall mediocre career. But to classify a player of this quality as mediocre would be mislabelling, so he was classified as excellent.
- Greg Oden was unable to have a meaningful NBA career due to injuries, playing essentially one season in the NBA. However, his talent was considered immense and it was almost certain that had he stayed healthy, he would have had a tremendous career. Therefore, the potential talent for this player was also classified as excellent.

As can be understood from the above examples, the assessment of potential talent was based more on the author's criterion, rather than on statistical data. We did not use a specific formula for evaluating a player's quality, nor did we attempt to create our own method. Inevitably this option will lead to data mislabeling, which will be more apparent in classes that report on lower-level talent.

However, the aim of this study was not to propose a method of evaluating a player's talent or quality with a computational method. Our goal was to propose a talent prediction method and that is what we focused on. The talent evaluation process, which was done for each player individually, was a time-consuming process, which can nevertheless be completed in a reasonable amount of time, as the size of the overall database allows it.

The categories (classes) in which this classification was made were the following five:

- **Not-ready:** Players that did not have the talent required to play in the NBA level and did not manage to have an impact on this level of competition
- **Lower level:** Players that were part of an NBA team roster, but they were limited-role players at best, with low playing time
- **Mediocre:** Players that had some decent seasons and had a significant role in the teams they played, but were not considered among the stars of the league
- **Very Good:** Players that had a good overall career but teams were not necessarily "built" around them, however they were essential parts of their teams
- **Excellent:** Players that had the talent to be an All-Star for one or multiple seasons, or were particularly good in a certain part of the game (rebounding for example)

The assessment of the players was made based on their whole career and not on particular season. The rational of this decision lies on the the multi-factor parameters that can affect a player's performance in a particular season. A player being on a team that is uncompetitive for one season (therefore even a player of average talent, has the opportunity to increase his averages), a good season by a player that didn't continue in subsequent years, are factors that are misleading in talent evaluation. Based on this reasoning we decided to disregard particular seasons of a player (either good or bad) and focus on the bigger picture of a player's career.

Can a face tell us anything about an NBA prospect?

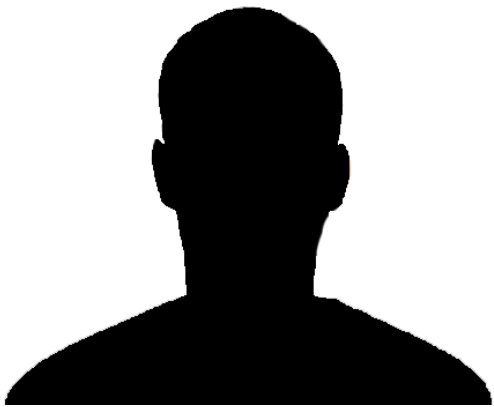

Figure 1: Example of the typical player image that was used in the database

### 3.3 Data Collection

The construction of the database was based on the following logic. The models should be trained exclusively on the athlete's facial features. Physical characteristics of an athlete's body were deliberately excluded from the analysis. An example of the typical image that used in our database, is provided in Fig.1. Also, as the aim of the study is to explore the potential of a player as a rookie, it was decided to use images of the concerned athlete as a rookie. The images of the athletes were accessed via web scraping. As a result of this, we were not able to have a consistency in the images we used to form our database. Under optimal conditions we would like the database to consist exclusively of photos which would have been taken with the same procedure, would have the same dimensions, the players would have the same pose, etc. Of course, this condition cannot be met, as the players of the last thirty years are examined. It is commonly understandable that as we go back in time the data becomes more and more inaccessible and the diversity in the way players are photographed increases. Therefore our database consists of highly differentiated images, a problem we tried to address through photo editing, to create a more "homogenized" set of images.

Table 1: Distribution of quality classes in the database

| Quality Class | Number of players |
|---|---|
| Not-ready | 607 |
| Lower level | 352 |
| Mediocre | 327 |
| Very Good | 237 |
| Excellent | 146 |

### 3.4 Database

The final dataset consists of data for 1669 rookies from 1990 to 2019. The distribution of classes is presented in Table.1. It is obvious that there is an uneven distribution of data, as the "Not-ready" class makes up almost 1/3 of the total database. Also, the "Excellent" class is considerably smaller than all the other classes. This picture of the database is rather expected and reflects reality. Most of the players in each draft class fail to stand up at the NBA level, let alone be good prospects. Of course this distribution of data will have an impact on the training of deep learning models, a problem that will be discussed later. The size of the resulting database is quite small compared to most projects were CNNs are deployed. However as we are dealing with a real-life problem it is another challenge we have to face.

### 3.5 CNN model development

As this study deals with a problem that is image-based, we chose to deploy *Transfer Learning*. *Transfer Learning* is one of the most popular options in image classification problems such as the one we are dealing with. The most well-known models have been trained for the needs of the ImageNet Challenge, which is a quite complex and demanding task. They have the ability to find patterns in image data that are related with emotions, diseases, etc [20, 21]. Therefore, we chose to use some of the most popular pre-trained models in this study as well.

The pre-trained we trained on our database were *ResNet-50*, *Xception*, *Inception-V3* and *VGG-16*. *VGG-16* model stacks together multiple convolutional, followed by a max-pooling layer. A 1x1 filter is set as last convolution layer in some layers, to induce non-linearity in the model, thus making it able to be deployed in more complex problems [22]. *ResNet-50* is a very deep CNN model consisting of 50 layers.The central idea of this model, is that it is built around the concept of *skip-connections*.

Skip-connections are deployed to connect the input directly to the output of the layer, while skipping a number of connections [23]. *Inception-V3* is making use of various filters of varied size in order to prevent overfitting (the phenomenon of a model overly adapting to the training data). Its main difference with other pre-trained models, is that it has a "wider" network architecture, consisting of multiple convolutional layers, that do not reach great depth. *V3* in particular requires less computational resources, in comparison to *V1* and *V2* so it was considered as the the optimal choice out of the Inception "family" of models [24]. The last model we used was *Xception*. It is wider known as an extreme version of Inception model. This results from the depthwise convolution it deploys, which is also followed by a point-wise convolution, which is not applied in all layers in order to be more efficient in terms of computational cost [25].

Table 2: Distribution of quality classes in the dataset that was used to train the pre-trained models

| Quality Class | Number of players |
|---|---|
| Not-ready | 300 |
| Lower level | 300 |
| Mediocre | 300 |
| Very Good | 230 |
| Excellent | 140 |

The small size of the resulting database prevented us from conducting several tests for each pre-trained model. For each model, more than 50 tests of different network architectures were trained. In the models that proved to be more successful, even more tests were conducted, reaching more than 100. Tests included the use of different cost functions, activation functions, number of training epochs, learning rate, optimizers, number of dense layers and the number of neurons in each dense layers. Also tests were made on the number of hidden layers of the pre-trained models that would be trained on the data.

Finally, to deal with the problem of unequal distribution of data in the classes it was chosen to truncate some classes to create more evenly distributed data. It was chosen to delete inputs from the classes that had more data in favour of the top talent classes. The resulting final dataset is presented in Table 2. We made this decision in order to avoid training the models on a particular class, that would probably had an effect on the quality of the predictions. Training sessions on the initial dataset would put us in the risk of producing CNN models that would be overly trained in players of lesser quality. Thus, it would make sense for the models to produce more predictions involving players of that class. Our goal is on the first level to produce reliable predictions for all classes, but on the second and equally important level to predict which players will have the biggest impact in the league. Therefore, our strategy was mainly driven around producing models that are more capable of producing predictions for a good talent.

## 4 Results

### 4.1 Results evaluation method

We used a 10-fold cross-validation process to evaluate the performance of the CNN models. In this evaluation process, the database was split into 10 equally sized folds. Nine folds are used to train the model and the remaining fold (10% of the database) was used as the unknown sample, where the model was called to make predictions. This process is eventually repeated for 10 times, (as the number of folds) until all the dataset inputs are used for evaluation.

The metric we used to evaluate the performance of the CNN models was Accuracy, which is the most common metric in image classification projects. However, as we are interested in producing predictions that are more driven towards evaluating high-talented player prospects, we are interested in one metric that would evaluate the quality of predictions for each distinct class. This metric is *Precision*, but we chose to rename it to be more understandable by people with no deep knowledge of DL metrics. This metric is more oriented towards evaluating class-specific predictions for the following reasons: If a model manages to achieve a decent accuracy rate overall but has very low performance in *Very Good* or *Excellent* classes, it has practically very low value for an NBA organization. By the same token, a model that does not achieve a high accuracy rate compared to other models, but manages to have higher success rates in the classes that refer to more talented players, then that model clearly has more value for the purposes of this study.

The name of the created metric was *Class-Prediction Quality* and is defined as follows:

$$\text{Class-Prediction Quality (\%)} = \frac{\text{Number of Successful Predictions for a distinct class}}{\text{Overall number of predictions for a distinct class}} \times 100 \quad (1)$$

This metric takes more into account the predictions that a model produces. If a model produces 30 predictions for *Very Good* class and manages to be accurate in 10 occasions, it should be considered as a well-performing model. Of course, this metric can be biased if the total number of predictions is significantly low, so we are taking into account the total number of predictions for the concerned class, as well.

Can a face tell us anything about an NBA prospect?

Table 3: Accuracy rates (%) of the best performing model of each group of pre-trained models in the unknown sample of images

| Pre-trained model type | Accuracy rate (%) |
|---|---|
| VGG-16 | 22.6 |
| Inception-V3 | 23.62 |
| ResNet-50 | 24.49 |
| Xception | 26.77 |

## 4.2 Evaluation of model performance

The accuracy rates in the unknown sample of images were significantly low, however this was an expected and desired result. Having high accuracy would be an alarming indicator that something went wrong in the training process or in the database configuration. The performance (accuracy rates) of the most successful models ranged between 20-25%. Next we will present the performance for each type of pre-trained model separately.

### 4.2.1 VGG-16

*VGG-16* models did not provide high precision rates in the unknown sample of images. Although, some of the models managed to achieve accuracy rates of 24%, this was the low-end performing model as far as *Class-Prediction Quality* is concerned. The performance of *VGG-16* models in the higher talent classes was rather low, with no network architecture achieving to overcome the threshold of 10% in *Very Good* and *Excellent* classes. In the classes that referred to lower level of talent, the models performed considerably better, with a model surpassing the mark of 25% accuracy for *Not-ready* and *Lower level* classes.

### 4.2.2 ResNet-50

The models of this group provided arguably the worst performance. The *Class-Prediction Quality* rates in two highest talent classes were almost zero in all the network architectures tested. These models provided decent performance in the other classes, recording *Class-Prediction Quality* rates above 25% in the concerned classes, however this was not the point of this research. As the results for this group of models were not at all encouraging, we chose to stop testing on more network architectures in order to focus on the best performing models. At this point it is worth noting that the failure of the first two models to successfully predict players with greater talent is a rather encouraging indicator. Finding talent is quite difficult even for human experts, so the fact that the models fail in this area may prove that the models simulate human behaviour and they have an equally difficult time finding high-level athletic talent.

### 4.2.3 Inception-V3

*Inception-V3* models managed to achieve decent accuracy rates, as well as decent *Class-Prediction Quality* scores in all classes. The best performing model from this group, managed to display and accuracy rate of over 20% in four classes (*Not-ready, Lower level, Mediocre, Excellent*) and over 18% in *Very Good* class. The best performing models had a rather low number of neurons in the dense layers (<100) and all these models deployed *Softmax* activation function. Although we tested the effect of deploying *Dropout* technique in the dense layers, we observed that its implementation had the opposite of the expected results.

### 4.2.4 Xception

The models of this group recorded the best performance with big difference from the other pre-trained models we used in the tests. Several models exceeded the 25 % threshold in accuracy rates. Also, there was uniformity in the distribution of predictions, as several models had *Class-Prediction Quality* scores of above (20%) in all classes. In fact, a small number of models managed to have *Class-Prediction Quality* rates above 25% in all classes, including the higher talent classes. This performance was achieved with a moderate number of neurons in the dense layers (100-500). It is worth noting here that in the models of this group, the most successful network architectures used 3 dense layers. In all these layers the *Dropout* technique was applied, which had a beneficial effect on the performance of the models. Specifically, it increased performance by 3-4% compared to models of identical network architecture in which *Dropout* was not implemented.

The model that achieved the highest performance in all classes will then be presented in more detail.

Can a face tell us anything about an NBA prospect?

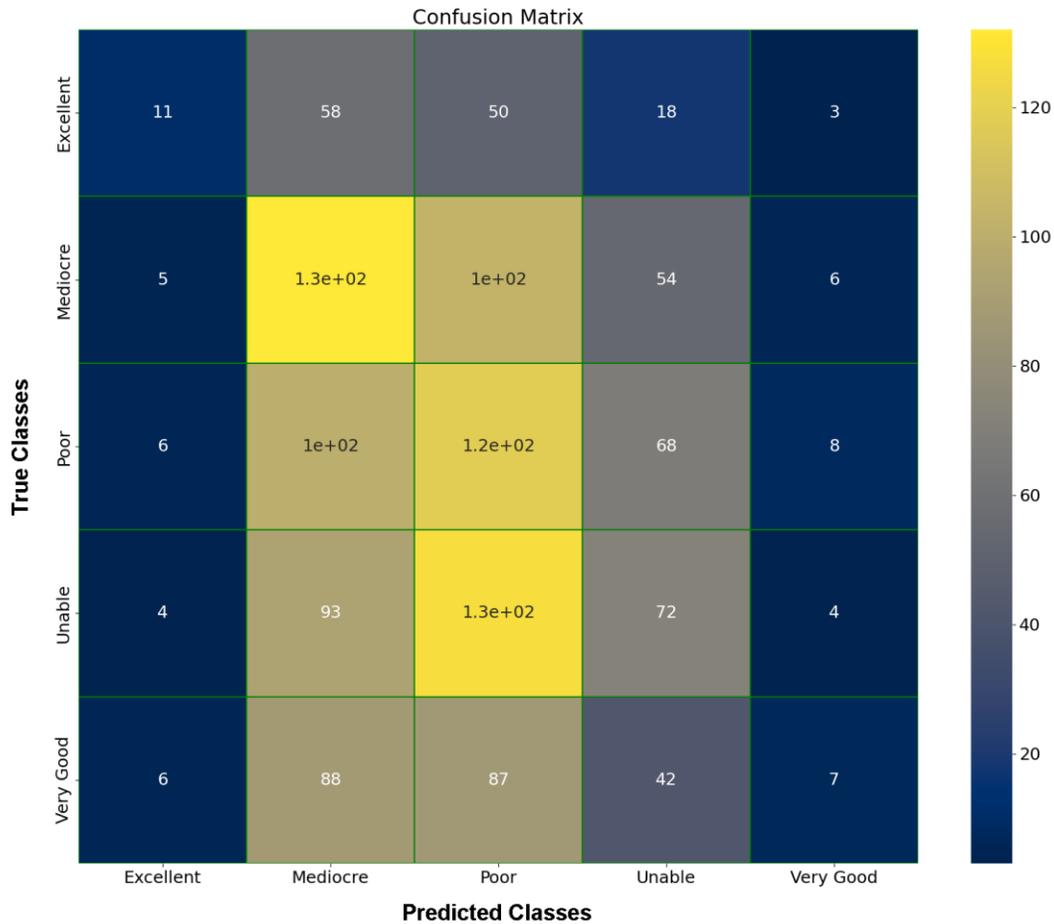

Figure 2: Confusion matrix of the predictions made by the best performing model

Table 4: Accuracy rate of best performing model, in each class

|  | Not-ready | Lower level | Mediocre | Very Good | Excellent |
|---|---|---|---|---|---|
| Class-Prediction Quality (%) | 28.35 | 24.33 | 28.03 | 25 | 34.38 |

### 4.3 Best performing model

The best performing model achieved an overall accuracy score of 26.77 % in all classes. This accuracy rate was among the highest in all the tests we conducted. However, the selection of this model as the best, was based in its performance in the classes of higher talent. The model performance by class is presented in the Table4. An overview of the predictions made by this model is also provided in Fig.2. As shown in the confusion matrix the model does not produce a large number of predictions for the *Very Good* and *Excellent* classes. The predictions produced are among the most accurate observed in the test cycles carried out, as shown by the decent rates of *Class-Prediction Quality*.

Regarding the Excellent class, the *Class-Prediction Quality* is quite high, approaching 35%. The number of total predictions produced (32) indicates that the performance is not due to statistical error. The number of total predictions produced (32) indicates that the performance is not due to statistical error. The same applies to the Very Good class where the *Class-Prediction Quality* is not at the same level (25%), but it is still a quite remarkable performance based on a reliable number of predictions (28). The number of predictions in the model is significantly lower compared to the other classes, a feature that may be desirable in the sports industry. Logically, teams would want a model that does not produce a large number of predictions, thus increasing the number of players that will be misjudged. We believe that a smaller number of predictions with the existing accuracy is much more manageable by teams.

Can a face tell us anything about an NBA prospect?

Finally, we found it interesting to deploy this model to generate predictions for this season's rookie players, and specifically for players coming out of the second round of the draft, where such a model would have more potential for use. The model ranked only one player in the top two talent classes, namely the Excellent class. This choice seems logical based on the fact that players selected lower in the draft have a lower chance of developing into league stars. The player selected by the model was EJ Liddell who was drafted at number 41 by the New Orleans Pelicans.

## 5 Conclusions

The performance of the models suggests that there is a potential correlation between players' talent and the characteristics of their foresight, This statement however, raises more questions than it answers. In the event that a correlation does exist, what are the characteristics that models see as associated with a person's predisposition to perform well in a particular sport? The present research is not able to provide a reliable answer to these questions. Future studies may be able to provide more data on the problem we faced here.

The results of this survey are quite encouraging, as in the high talent classes the accuracy rates are high, subject to proportions of course. A first question is obvious: could these results have been better? As the images of the players used were in many cases of low resolution, a first idea is to create a database of higher resolution images. Then as there is a large heterogeneity in the images of the players, a possible improvement of the accuracy rates could be based on a dataset consisting of images of a certain posing-style.

Finally, we would like to dwell once again on the nature of this work. This research would be optimally seen as an exercise on the unexplored potential of CNNs. To appear confident that the results of this paper represent a certain conclusion would be unrealistic.

However, we do not consider it unlikely that in the future NBA teams will photograph rookie players as part of their evaluation during their interview process. The methodology developed in this paper may be part of the talent analysis tools for NBA teams in the future.

### Acknowledgments

Most of the work dedicated to conducting this research was done under quarantine conditions for COVID-19 in the spring of 2020. Therefore, the authors of this research would like to dedicate this work to the relatives of the victims of this disease, who experienced this unprecedented situation in the worst possible way.